# Learning Crisp Boundaries Using Deep Refinement Network and Adaptive Weighting Loss

Yi-Jun Cao, Chuan Lin, and Yong-Jie Li, *Senior Member, IEEE*

*Abstract*—Significant progress has been made in boundary detection with the help of convolutional neural networks. Recent boundary detection models not only focus on real object boundary detection but also "crisp" boundaries (precisely localized along the object's contour). There are two methods to evaluate crisp boundary performance. One uses more strict tolerance to measure the distance between the ground truth and the detected contour. The other focuses on evaluating the contour map without any postprocessing. In this study, we analyze both methods and conclude that both methods are two aspects of crisp contour evaluation. Accordingly, we propose a novel network named deep refinement network (DRNet) that stacks multiple refinement modules to achieve richer feature representation and a novel loss function, which combines cross-entropy and dice loss through effective adaptive fusion. Experimental results demonstrated that we achieve state-of-the-art performance for several available datasets.

*Index Terms*—Contour detection, Decode Network, Deep Refinement Network, Multi-Scale Integration.

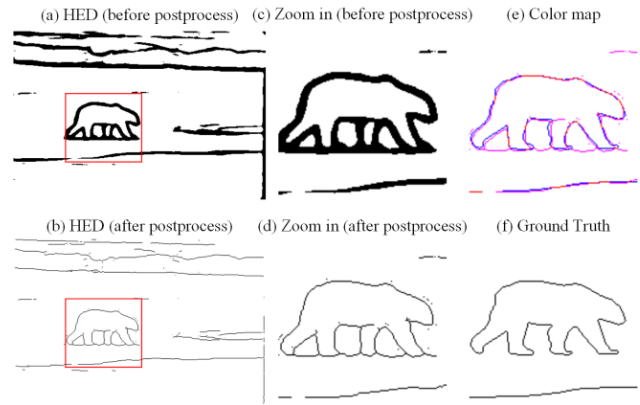

Fig. 1. Visualization of edge maps from HED with and without post-processing. (a) and (b) show the HED result before and after post-processing (NMS and morphological thinning) with a fixed threshold of t = 0.5. (c) and (d) show the corresponding zoomed images. (e) Color map, where the red, blue and purple pixels indicate $E_{GT} \cap E_a \cap E_b$, $E_{GT} \cap D_a \cap E_b$, and $D_{GT} \cap E_a \cap E_b$, respectively. Here $E$ and $D$ are the contour and background union, respectively, and subscripts $GT$, $a$, and $b$ indicate the ground truth, and HED output after and before post-processing, respectively. (f) Ground truth image.

## 1 INTRODUCTION

**B**OUNDARY detection plays a fundamental role in image and video processing applications such as local surface descriptors [1] and image retrieval [2, 3]. In these applications, boundary detection is usually considered a low-level technique and normally used as a shape or boundary constraint to help high-level image tasks improve their performance.

In the past few years, convolutional neural networks (CNNs) have become popular in the computer vision field and have helped substantially improve many tasks, including image classification [4, 5], object detection [6], and semantic segmentation [7]. The specific architectures of CNNs can be exploited for high-performance contour detection exploits CNNs architectures for higher performance. Many well-known CNN-based methods such as DeepContour [8], HED [9], RCF [10], and CED [11] have shown remarkable improvements beyond the state-of-the-art F-score of benchmarks methods such as BSDS500 [12] and NYUDv2 [13].

Although some state-of-the-art methods achieved human-like performance on standard datasets such as BSDS500 [12], contour detection is still a challenging problem for extracting edges with a high level of detailed. The main limitation of CNN-based methods is that they extract highly "correct", yet less "crisp" edges (i.e., are not well localized) [11]. As shown in Fig. 1(e), the red and blue pixels indicate $E_{GT} \cap E_a \cap E_b$ and $E_{GT} \cap E_a \cap E_b$, respectively, with a large tolerance, where $E$ is the contour union and subscripts $GT$, $a$, and $b$ are the ground truth, HED output after and before post-processing, respectively. The "correct" contour (blue edges) are adequate for describing the object, but not sufficiently "crisp" (red edges). This issue is deeply rooted in modern CNN architecture [4], where the pooling layers lead to blurred output of edges and a fully convolutional architecture encourages similar responses of neighboring pixels, thus failing to produce a thin edge map. In addition to contour detection, semantic segmentation and target detection tasks are also focused on solving fine positioning problems.

This work was supported by the National Natural Science Foundation of China (Grant No. 61866002), Guangxi Natural Science Foundation (Grant No. 2018GXNSFAA138122 and Grant No. 2015GXNSFAA139293), Innovation Project of Guangxi Graduate Education (Grant No. YCSW2018203).

Yi-Jun Cao and Chuan Lin are with the College of Electric and Information Engineering, Guangxi University of Science and Technology, Liuzhou 545006, China (e-mail: yijuncao@gmail.com; chuanlin@gxust.edu.cn). Corresponding author: Chuan Lin.
Yong-Jie Li is with the School of Life Science and Technology, University of Electronic Science and Technology of China, Chengdu 610054, China (email: liyj@uestc.edu.cn).

To solve the challenging problem of crisp edge detection using CNNs, Wang et al. [11] proposed a CED model with a backward-refining decoding part fusing the feature map with intermediate features along the forward-propagating encoding part. With the help of sub-pixel convolution [14] for upsampling, CED achieves quite crisp boundaries. Starting from the loss function, Deng et al. [15] introduced a novel loss function with fusing weighted cross-entropy and dice coefficient for boundary detection.

Inspired by the backward-refining decoding architecture, we propose a novel deep refinement network (DRNet) which expands previous refinement architecture in a hierarchical manner that deepens the refinement network to extract detailed features. Specifically, some refinement modules explicitly exploit information available along the side outputs of the encoding process, and the other refinement modules try to refuse the output features. In addition, considering the multi-annotation problem in the contour detection task, such as in a BSDS500 dataset including 5–10 annotations per image, we simply revised the original weighted cross-entropy and dice coefficient to make full use of the label's information. Finally, the two losses are weighted adaptively to observe which one performs better and has fewer hyperparameters compared with setting weights manually.

Beyond the same goal of crisp boundary detection, different studies differently evaluate crispness with a boundary map. Wang et al. [11], followed the evaluation method proposed in [16], take a tighter matching distance, while Deng et al. [15] consider that directly evaluating boundary maps without standard non-maximal suppression (NMS) is a simpler yet effective for the same purpose. There are important differences between these the two methods. If a pixel is detected using a small tolerance, it is not necessarily detected using a large tolerance, e.g. compare the red and blue pixels in Fig. 1(e). Furthermore, if a pixel is true positive on the original HED output, the pixel is not necessarily detected on the map after post-processing, e.g. purple pixels in Fig. 1(e) corresponding to $D_{GT} \cap E_a \cap E_b$, where $D$ is the background union. The evaluation method with a short matching distance mainly focuses on finding highly localized edges, called localness, while the other method simply evaluates the model output without any other process, called thickness. Thus, in this study, we use localness and thickness to evaluate the crispness of boundaries.

In most cases, improving one of these metrics results in a decrease in the performance of the other, which highlights the challenge of simultaneously improving the localness and thickness. Previous studies try to solve this via either the CNN architecture, e.g., CED [11] or the loss function [15]. However, these methods were unable to improve localness and thickness simultaneously. To address this, we used our DRNet and developed a loss function with the aim of achieving state-of-the-art performance. Detailed ablation experiments were performed to evaluate the effect of each component on the localness and thickness boundaries. Our contributions are summarized into three parts.

1. We propose a novel network DRNet comprising multiple layers, which each contain multiple refinement modules. DRNet can integrate hierarchical features and improve localness performances.
2. We propose an effective loss function involving an improved cross-entropy and Dice loss, which were adaptively combined to automatically update the weights for convergence. The proposed loss is a simple extension to multi-loss fusion and improves performance.
3. Ablation experiments were conducted to determine which component helps improve crisp boundary detection. Our method outperforms previous state-of-the-art methods using the BSDS500 and NYUDv2 datasets.

The remainder of this paper is organized as follows. Section 2 presents related work. Section 3 describes our CNN architecture and loss function. In Section 4, we evaluate localness and thickness edges, describe the detailed ablation study, analyze the influence of each component, and compare our method to current state-of-the-art boundary detection methods. Finally, in Section 5, we discuss our work with respect to possible future directions.

## 2 RELATED WORKS

Early contour detection approaches were focused on finding local discontinuities, normally brightness, in image features. Prewitt [17] operators detect edges by convolving a grayscale image with local derivative filters. Marr and Hildreth [18] used zero crossings of the Laplacian of Gaussian operator to detect edges. The Canny detector [19] also computes the gradient magnitude in the brightness channel, adding post-processing steps, including NMS and hysteresis thresholding.

More recent approaches use multiple image features such as color and texture information and apply biologically motivated methods [20-22] or learning techniques [23-29]. Martin *et al.* [23] built a statistical framework for difference of brightness (BG), color (CG), and texture (TG) channels and used these local cues as inputs for a logistic regression classifier to predict the probability of boundary (Pb). For integrating multiple scales of information, Ren *et al.* [25] used local boundary cues, including contrast, localization, and relative contrast. To fully utilize global visual information, Arbelaez *et al.* [12] proposed a global Pb (gPb) algorithm that extracted contours from global information by using normalized cuts to combine the above-mentioned local cues into a globalization framework. Ren *et al.* [26] used sparse code gradients (SCG) to extract salient contours from sparse codes in oriented local neighborhoods. Lim *et al.* [27] proposed a fast and accurate edge detector for both learning and detecting local contour-based representations of mid-level features called sketch tokens. Khoreva *et al.* [30] introduced the problem of weakly supervised object specific boundary detection and suggested that good performance can be obtained on many datasets using only weak supervision (*i.e.*, leveraging bounding box detection annotations without the need for instance-wise object boundary annotations). Dollar et al. [31] used random decision forests to represent the structure of local image patches. By inputting color and gradient features, the structured forests output high-quality edges. However, all such methods were developed based on handcrafted features, which are limited in terms of their ability to represent high-level information for semantically meaningful edge detection.

In recent years, convolutional neural networks (CNNs) have been widely used in computer vision and machine learning. Ganin *et al.* [32] proposed $N^4$-Fields, which uses a deep architecture to extract features of image patches. They approached contour detection as a multiclass classification task by matching the extracted features to predefined ground-truth features. Bertasius *et al.* [33] made use of features generated by pretrained CNNs to regress and classify the contours. They proved that object-level information provides powerful cues for the contour prediction. Shen *et al.* [8] learned deep features using mid-level information. These methods only used CNNs as a local feature extractor and classifier rather than using global image information to build end-to-end training. Inspired by FCN [19], Xie and Tu [9] developed an end-to-end CNN, called HED, to boost the efficiency and accuracy of contour detection using convolutional feature maps and a novel loss function. HED connects its side output layers, which are composed of one *conv* layer with kernel size 1, one *deconv* layer and one softmax layer, to the last *conv* layer of each stage in the VGG16 network. Based on this architecture, Kokkinos [34] built a multi-scale HED algorithm and improved the results by tuning the loss function and adding globalization. Liu et al. [10] added side output layers to HED to extract richer convolutional features. In addition, Yang et al. [35] proposed an encoder–decoder architecture for contour detection task. Maninis et al. [36] proposed modeling the orientation of edges. Xu et al. [37] introduced a hierarchical model to extract multi-scale features and a gated conditional random field to fuse them. He et al. [38] proposed a bi-directional model to further fuse multi-scale information. However, these results tend to emphasize the "correctness" of the edges by selecting an optimistic matching distance and overlook the "crispness" of edges.

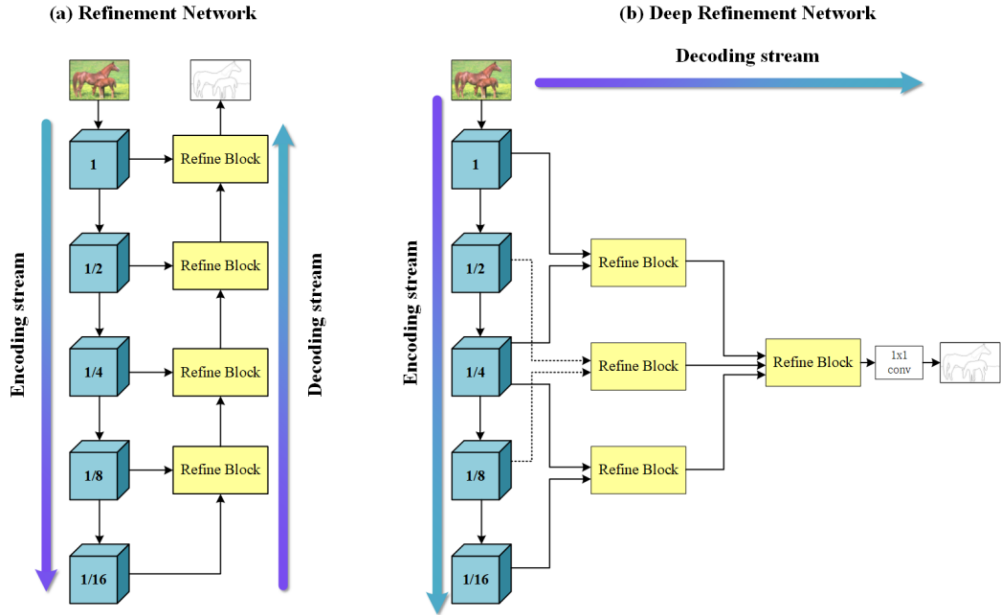

Fig. 2. Comparison of our DRNet and a previous refinement network. (a) Common refinement architectures. (b) DRNet.

To improve the "crispness" of edges further, Wang et al. [11] combined the refinement scheme with sub-pixel convolution [14] to produce a novel architecture that was specifically designed for learning crisp edge detection. Deng et al. [15] introduced a novel loss for boundary detection, which is very effective for classifying imbalanced data, and enabled the CNNs to produce crisp boundaries.

Our method is motivated by the works of Wang et al. [11] and Deng et al. [15]. Both proposed an end-to-end bottom–up/top–down architecture for learning crisp boundaries, while Deng et al. [15] introduced the dice coefficient to obtain thick edges. We share the same goal of designing a crisp edge detector, yet our method further improves the performance. Finally, our method is inspired by Kendall et al. [39], wherein they proposed a group of learnable weighted variables for multi-task learning.

## 3 PROPOSED METHOD

In this section, we describe the details of the proposed method. We first describe the proposed DRNet and each component used in building this architecture. We then describe the adaptive weight loss developed by fusing the improved cross-entropy and dice loss.

### 3.1 Network architecture

Generic refinement architecture (e.g., [11] and [15]) includes a top–down encode network, similar to the instance on the left side of Fig. 2(a), and a down–top decode network that aims to fuse multi-resolution side outputs of encode parts. Previous studies have shown that this architecture excels at capturing hierarchical features and is able to fuse them at different stages to generate semantically meaningful contours. An effective structure allows features with different resolutions to be blended. Hence, our DRNet structure, as shown in Fig. 2(b), horizontally expands the depth of the network to extract richer and more complex feature maps to increase the generality of the model.

The core of DRNet consists of remix refinement modules. We integrate multi-level features for high-resolution prediction in end-to-end computer vision tasks. Compared to a traditional refinement network that combines intermediate features along the forward-propagating pathway, DRNet fuses the refinement outputs, resulting in more accurate learning features. The VGG-

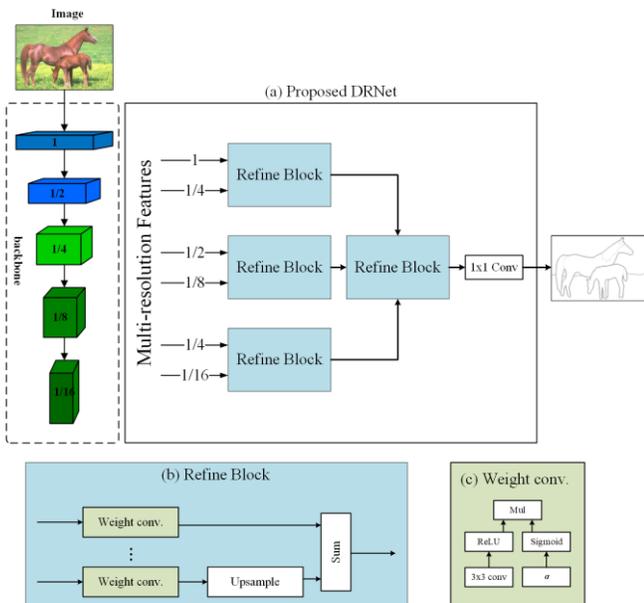

Fig. 3. Detailed DRNet architecture. (a) DRNet fusing multi-resolution features. (b) Refine block in our architectures. (c) Proposed weight convolution layer.

16 or ResNet was used as backbone, where we stack the 'refactored' structure of the refinement module to recover the resolution of the features. The detailed architecture is shown in Fig. 3. The following development of the network was performed to make it suitable for edge detection.

**Connections.** Two types of connection methods exist in the refine blocks: skip and adjacent connections. The former receives feature maps with four-time resolution difference. For example, in Fig. 3(a) the right module receives (1, 1/4), (1/2, 1/8) and (1/4, 1/16) size feature maps as input and then up-samples the lower one for fusion. The adjacent connection receives three adjacent inputs, as shown in Fig. 3(a), as the right refine block. The block fuses the different resolution map by up-sampling every low-resolution part to the highest one.

**Refine block.** Fig. 3(b) shows the detailed refine block in the proposed architecture. The refinement module includes up-sample, point-wise sum, and weight convolution layers. As the feature maps in each stage of backbone have different resolutions, we use bilinear up-sampling to up-sample the low-resolution feature maps into a high-resolution one. In addition, we use an element-wise add layer to fuse multiple inputs rather than a $1 \times 1$ convolution layer after concatenating multiple inputs because the element-wise add layer is a trade-off between performance and the number of training parameters. The output channel is set to the dimensions of the smallest one in the same refine block. After the last refine block, a $1 \times 1$ convolution layer compresses the features into one channel.

**Weight convolution.** Different resolutions will affect feature proportion of the final output boundary. Thus, we designed a weight convolution block for automatically balancing each feature. As shown in Fig. 3(c), the weight convolution layer includes two streams, one applying a normal $3 \times 3$ convolution following ReLU layer, and another applying a sigmoid function activated trainable parameter $\alpha$. The feature map obtained by the first stream is automatically weighted by the second one. In our experiment, the weight convolution layer should improve the performances compared with the traditional convolution layer and has less computational cost.

Our network is simple yet very effective for edge detection. During the forward pass, these refine blocks of first level convey and decode the multi-resolution features and then transmit them into separate refine blocks for further refining. Compared to the original structure, our network has the advantage of using similar parameters to achieve better performance.

### 3.2 Adaptive weigh loss for boundary detection

The loss function is the most important component in end-to-end CNNs as the quality of prediction is most affected by its loss. In CNN-based boundary detection, loss design generally considers: 1) positive and negative sample trade-off; 2) application of multi-annotation information in binary classification; 3) multiple losses combination; and 4) contour performance development. Considering at the last three points, we applied weighted cross-entropy and the dice coefficient loss (Section 3.2.1) to make full use of multi-annotation information, thereby improving performance. Then, considering the balance of cross-entropy and dice loss, we propose a novel method to combine these losses without hand-specific weights and obtain good performance.

### 3.2.1 Improved cross-entropy and dice coefficient loss

We start by considering the weight cross-entropy proposed in HED[9]. It is well known that the cross-entropy loss is used to solve binary classification. However, the edge/non-edge pixels have a highly imbalanced distribution as most of pixels are non-edge; thus, the direct use of cross-entropy loss would fail to train the network. To address this issue, HED uses weighted cross-entropy loss:

$$\text{CE}(P, L) = -\beta \sum_{i \subset L_+} \log p_i - (1-\beta) \sum_{i \subset L_-} \log(1-p_i) \quad (1)$$

where $L_+$ and $L_-$ denote the edge pixel set and non-edge pixel sets, respectively. In addition, $\beta = |L_-|/|L|$, $1 - \beta = |L_+|/|L|$ and $p_i$ is the CNN output processed by a sigmoid function at pixel $i$. Using weights $\beta$ and $1 - \beta$, this solution help CNNs better train the network.

HED adopts quite a brute-force solution for the multiple annotator problem. It only assigns a positive label to a pixel if it is labeled as positive by at least three annotators (i.e., all other pixels are labeled negative). Subsequently, RCF [10] considers a pixel as positive only if it is labeled by more than 50% of the annotators, where pixels without any positive label are treated as negative samples. Otherwise, if a pixel is marked by fewer than 50% of the annotators, this pixel is ignored during training (by blocking their gradients).

However, the above-mentioned methods seem a bit too aggressive. Although such human-labeled edge maps (5–10) share a little consistency, some potentially information could be discarded. Thus, we convert the problem of assigning positive and negative labels to samples into a weighting problem. Thus, we first transform multiple annotation averaging into a weight map:

$$W = \frac{1}{n} \sum_{i=1}^{n} L^i \quad (2)$$

where $n$ is the number of annotators for each image and $L^i$ indicates the *i-th* annotation map. $W$ is the probability label map formed by averaging all ground truth, which ranges from 0 to 1. We consider a pixel positive if $W > 0$, i.e., it is labeled positive by at least one annotator.

Then, each positive labeled pixel is weighted using $W$. The new soft cross-entropy (SCE) defined by:

$$\text{SCE}(P, W) = -\frac{\beta}{|Y_+|}\sum_{i \subset \{W>0\}} W_i * \log p_i - \frac{(1-\beta)}{|Y_-|}\sum_{i \subset \{W=0\}} \log(1-p_i) \quad (3)$$

where $\{W > 0\}$ denotes the positively label set and $\{W = 0\}$ is the non-edge pixel set. Each positive pixel $i$ is weighted using $W$. We also normalized positive and negative loss to avoid the imbalances using range of the value with the adaptive weighting method (Section 4.2.2). Compared with (1), the proposed loss is simple yet very effective for edge detection.

A distinct characteristic of the edge map is that it has highly biased data because the vast majority of the pixels are non-edges, which makes learning to find crisp edges challenging. To address this, Deng et al. [15] introduced dice coefficient loss for learning crisp boundaries:

$$\text{D}(P, L) = \frac{\sum_i^N p_i^2 + \sum_i^N l_i^2}{2\sum_i^N p_i l_i} \quad (4)$$

where $p_i$ and $l_i$ denote the *i-th* pixel on the prediction map $P$ and the label $L$, respectively. $N$ denotes the total number of input images. However, this loss does not consider multiple annotation issues. Thus, we simply revised it to suit different label maps and named soft dice loss (SD) as follow.

$$\text{SD}(P, W) = \frac{\sum_i^N p_i^2 + \sum_i^N W_i^2 + \varepsilon}{2\sum_i^N p_i W_i + \varepsilon} \quad (5)$$

Here, we simply substitute ground truth map $L$ by $W$. $\varepsilon$ is introduced to avoid "not a number" errors and was set to a value of $1\times10^{-6}$.

*3.2.2 Adaptive fusion*

To achieve a better performance, the SCE and SD can be combined as one loss function, which is given by

$$L(P, W) = \kappa \text{SCE}(P, W) + \tau \text{SD}(P, W), \quad (6)$$

where $\text{SCE}(P, W)$ is the improved cross-entropy loss shown in Equation 3; $\text{SD}(P, W)$ is the improved Dice loss shown in Equation 5; and $\kappa$ and $\tau$ are the parameters that control the influence of the SCE and SD losses, respectively. The uncertain parameters $\kappa$ and $\tau$ control the relative confidence between the losses. Although we can assign them manually, it is difficult to optimize these values to a balance point to enhance performance. Inspired by a previous study [39], which proposed the homoscedastic uncertainty as a basis for the weighting losses in a multi-task learning problem. We modified the proposed theory for combining multiple losses in a boundary detection task problem. Specifically, the final adaptive weight loss (AWL) function is defined as:

$$L_{final}(P, W) = \frac{1}{\kappa^2}\text{SCE}(P, W) + \frac{\zeta}{\tau^2}\text{SD}(P, W) + \log(1 + \kappa\tau) \quad (7)$$

The parameters $\kappa$ and $\tau$ are trainable, meaning that we only need to provide an initial value and then update it with the training of the network. The parameter $\zeta$ is fixed, and is often used to fuzzy balance different cost functions to an order of magnitude. The auxiliary term, $\log(1 + \kappa\tau)$ can be regarded as a weight learning term. In training, large values of $\kappa$ and $\tau$ decrease the contribution of the loss term $\frac{1}{\kappa^2}\text{SCE}(P, W) + \frac{\zeta}{\tau^2}\text{SD}(P, W)$, whereas small values of $\kappa$ and $\tau$ increase its contribution. The parameters $\kappa$ and $\tau$ are regulated by the last term. During training, both terms mutually interact and restrict each other, and thus, automatic balancing occurs. More importantly, this type of method has good extendibility, which means that Eq. (7) simply be revise to combine three or more loss terms.

In practice, we set the auxiliary term as $\log(1 + \kappa\tau)$ rather than $\log(\kappa\tau)$ due to its numerical stability. The parameter $\zeta$ is fixed during training to ensure that the output value of $\text{SCE}(P, W)$ and $\text{SD}(P, W)$ are within a certain range. The experiments in the next section show that this parameter is not strict, providing that the difference between different terms, such as $\text{SCE}(P, W)$ and $\text{SD}(P, W)$, does not exceed two orders of magnitude.

*3.3 Multi-scale contour detection*

We construct image pyramids to detect multi-scale contours, as used in previous studies [10, 11]. Specifically, an image is resized to construct an image pyramid, and each of these images are separately input to our DRNet. Then, all the resulting edge probability maps are resized to the original image size using bilinear interpolation. Finally, these maps are averaged to obtain a final prediction map. Considering the trade-off between accuracy and speed, we use three scales, namely, 0.5, 1.0, and 2.0, in this study.

4 EXPERIMENTS

*4.1 Implementation details*

**Hyperparameters.** The network was implemented using the publicly available PyTorch [40], which is well-known in this community. For training, the hyperparameters of our model included: mini-batch size (10), global learning rate (0.01), learning rate decay (0.1), weight decay ($1\times10^{-4}$). SGD was used and the momentum was set to 0.9. The trainable parameters $\kappa$ and $\tau$ were both initialized to 1.

**Data augmentation**. Data augmentation is an effective way to boost performance when the amount of training data is limited. First, the image-label pairs (0.75–1.25) were randomly scaled. Then, on the BSDS500 dataset [12], the pairs were rotated in 16 different angles and flipped the image at each angle. Finally, we flipped the cropped images, which resulted in an augmented training set from 200 images to more than 100000 images. On the NYUD-v2 dataset [13], the images were rotated with corresponding annotations at four different angles (0, 90, 180, and 270°) and flipped them at each angle. On MultiCue dataset [42], we randomly flip and crop the image to $500 \times 500$.

**Dataset.** We used BSDS500 [12], NYUD [13] and MultiCue [42] datasets for our experiments. BSDS500 [12] is a widely used dataset in edge detection. It is composed of 200 training, 100 validation, and 200 test images, and each image is labeled by multiple annotators. We used the training and validation sets for fine-tuning, and the test set for evaluation. Inspired by previous studies [10, 11, 34], we mixed the augmentation data of BSDS500 with the flipped PASCAL VOC Context dataset

[43] as the training data. During evaluation, standard NMS [31] was applied to thin the detected edges.

The NYUD [13] dataset is composed of 1449 densely labeled pairs of aligned RGB and depth images. Gupta et al. [44] split the NYUD dataset into 381 training, 414 validation, and 654 test images. We followed their settings and trained our network using the training and validation sets in full resolution as in HED [9]. We used the depth information using HHA [45], in which the depth information is encoded into three channels: horizontal disparity, height above the ground, and angle with gravity. Thus, the HHA features can be represented as a color image. Then, two models for the RGB images and HHA feature images were trained separately. During testing, the final edge predictions were defined by averaging the outputs of the RGB model and HHA model. During evaluation, we increased the localization tolerance, from *maxDist* = 0.0075 to 0.011, as in [10], because the images in NYUD dataset are larger than those in BSDS500 dataset.

The MultiCue dataset [42] is composes of short binocular video sequences of 100 challenging natural scenes captured by a stereo camera. The last frame of the left images is labeled for two kind of annotations, object boundaries and low-level edges. They strictly defined boundary and edge according to visual perception at different stages. Thus, boundaries are the boundary pixels of meaningful objects, and edges are abrupt pixels at which the luminance, color or stereo change sharply. According to previous study [9], we randomly split the dataset into 80 images for training and 20 for testing, and give the average scores of three independent trials as the final results.

*4.2 Evaluation of crisp boundaries*

As per previous studies [11, 15], good boundary detections must be both "correct" (detects real object boundaries) and "crisp" (precisely localized along the object's contour). Additionally, directly learning a thick boundary without any other post-processing is an important ability of modern CNNs as well. However, the standard evaluation benchmark [23] does not distinguish between these criteria. According to previous work, we used three criteria to evaluate correct and crisp boundaries.

The first criterion is the standard evaluation benchmark evaluating boundaries [23], which test the output images with a tolerance of $d_0 = 4.3$ pixels (corresponding to the parameter *maxDist = 0.0075* for BSDS500) after standard post-processing (NMS and morphological thinning). This default setting can evaluate the correctness of the boundaries; however, it cannot distinguish whether or not an algorithm captures the details of the edges. To evaluate the localness of the edges, we decreased the tolerance to $d = d_0/4$ and used standard post-processes for testing the localness contours, as in [11, 15]. The thickness of the edges was measured with the default tolerance $d$ without standard post- processing, as in [23].

The prediction accuracy was evaluated via three standard measures: fixed contour threshold (ODS), per-image best threshold (OIS), and average precision (AP). In the ablation study, we only used ODS to verify the three criteria (correctness, localness, and thickness), because the trends of ODS, OIS, and AP are similar. For conciseness, we use the notations, ODS-C, ODS-L, and ODS-T, to indicate ODS for correctness, localness, and thickness, respectively.

*4.3 Ablation study*

This section presents the ablation analysis used to assess the performance of all subsystems of our method. We trained using the BSDS *train and val* set and evaluated the BSDS *test* set. We report the ODS-C, ODS-L, and ODS-T for performance evaluation and named the proposed method with the VGG16 backbone as deep refinement contour (DRC).

Table 1. Ablation studies of the loss function.

| Method | SCE | SD | ODS-C | ODS-L | ODS-T |
|---|---|---|---|---|---|
| HED | × | × | 0.780 | 0.548 | 0.583 |
| | √ | × | 0.795 | 0.560 | 0.613 |
| DRC | × | × | 0.796 | 0.581 | 0.633 |
| | √ | × | **0.799** | 0.595 | 0.674 |
| | × | √ | 0.711 | 0.514 | 0.634 |
| | √ | √ | 0.798 | **0.598** | **0.681** |

Table 2. Comparison of loss weights that were selected manually or using the proposed AWL method.

| Loss Weights | | Performances | | |
|---|---|---|---|---|
| κ | τ | ODS-C | ODS-L | ODS-T |
| 0 | 1 | 0.711 | 0.514 | 0.634 |
| 0.25 | 1.75 | 0.770 | 0.563 | 0.658 |
| 0.5 | 1.5 | 0.783 | 0.578 | 0.673 |
| 0.75 | 1.25 | 0.790 | 0.586 | 0.680 |
| 1 | 1 | 0.798 | 0.598 | 0.681 |
| 1.25 | 0.75 | 0.796 | 0.592 | 0.680 |
| 1.5 | 0.5 | 0.789 | 0.588 | 0.684 |
| 1.75 | 0.25 | 0.791 | 0.589 | 0.682 |
| 1 | 0 | 0.799 | 0.595 | 0.674 |
| Learned weights ($\zeta = 0.01$) | | **0.802** | 0.600 | 0.694 |
| Learned weights ($\zeta = 0.1$) | | **0.802** | **0.601** | **0.697** |
| Learned weights ($\zeta = 1$) | | **0.802** | 0.599 | 0.694 |
| Learned weights ($\zeta = 10$) | | 0.796 | 0.586 | 0.665 |

**Loss function.** First, we tested the performance of SCE and SD. Table 1 shows that the proposed SCE loss outperformed the original version. Furthermore, we combined the SD loss directly, as in Eq. (6). In general, combining SCE and SD improved ODS-L and ODS-T performances, whereas independently adapting the SD loss to train the network resulted in poor results. Fig. 4 shows contour maps using different loss combinations without any post-processing. We can observe that the proposed SCE loss extracted more details and increased the crispness of the contour compared to other methods, while the use of SD loss for training achieved a better edge thickness at the price of losing many useful details. We compared the AWL with the manual selection of weights. Table 2 presents the models trained on different weight combinations, where the combined loss performed better than the individual losses. However, determining these optimal weightings is expensive and increasingly difficult for large models with numerous losses. The last four rows in Table 2 show that the quantitative result of the proposed AWL with different $\zeta$ values were better than the those of the manually selected weights. Although the increase is not large, the role of AWL will become more apparent as more cost functions are combined. Note that the range of the two losses (SD and SCE) was similar for $\zeta = 0.1$. The experiments demonstrated that the proposed AWL is very robust. When the range of losses differed by an order of magnitude (i.e., $\zeta = 0.01$ or $\zeta = 1$), the training result was still acceptable; however, when it exceeded two orders of magnitude (i.e., $\zeta = 10$), the effect was significantly reduced. Thus, $\zeta = 0.1$ was used in further experiments.

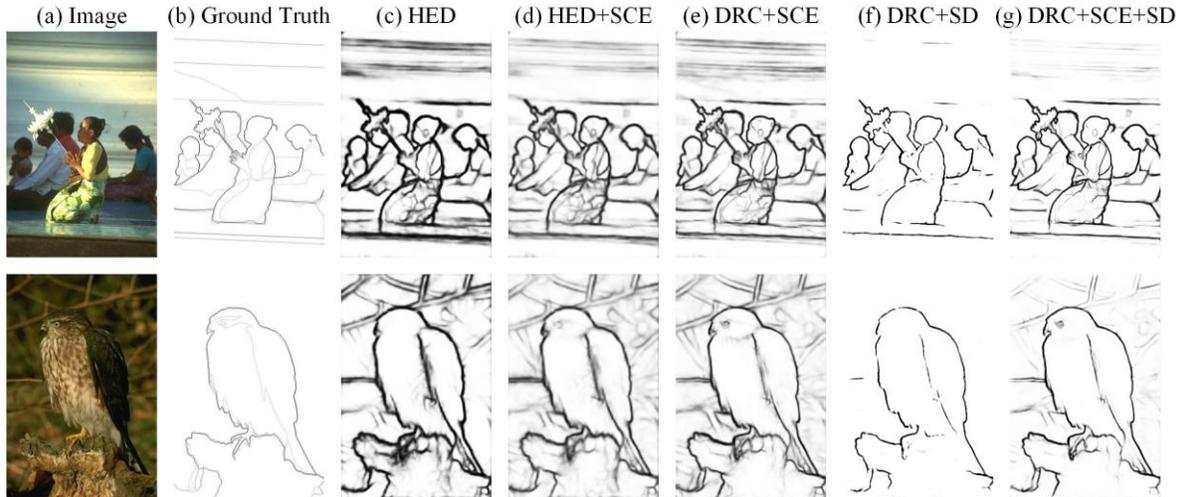

Fig. 4. Illustration of the qualitative results of the ablation study. (a) Input images in the BSDS500 dataset; (b) ground truth; (c, d) predictions of HED and HED with proposed SCE loss, respectively; (e, f, g) predictions of the proposed methods DRC with SCE, SD, and SCE+SD loss, respectively.

Table 3. Results of the ablation studies of the network and PASCAL VOC for mixed training.

| Method | MS | VOC | ODS-C | ODS-L | ODS-T |
|---|---|---|---|---|---|
| DRC-w/o-W.Conv. | × | × | 0.796 | 0.590 | 0.656 |
|  | × | × | 0.802 | 0.601 | **0.697** |
| DRC | √ | × | 0.808 | 0.604 | 0.672 |
|  | × | √ | 0.806 | 0.614 | 0.645 |
|  | √ | √ | 0.817 | 0.623 | 0.644 |
| DRC-Resnet50 | × | √ | 0.813 | 0.605 | 0.660 |
| DRC-Resnet50 | √ | √ | 0.819 | **0.631** | 0.655 |
| DRC-Resnet101 | × | √ | 0.812 | 0.603 | 0.657 |
| DRC-Resnet101 | √ | √ | **0.821** | **0.631** | 0.657 |

Table 4. Quantitative comparison of boundaries ($F_b$ [23]) on BSDS500 dataset.

| Method | ODS | OIS | AP |
|---|---|---|---|
| Canny [19] | 0.611 | 0.676 | 0.520 |
| gPb [12] | 0.729 | 0.755 | 0.745 |
| SE [31] | 0.743 | 0.764 | 0.800 |
| DeepContour [8] | 0.757 | 0.776 | 0.790 |
| DeepEdge [33] | 0.753 | 0.772 | 0.787 |
| HED [9] | 0.788 | 0.808 | 0.840 |
| AMH-Resnet50 [37] | 0.798 | 0.829 | 0.769 |
| COB [36] | 0.793 | 0.819 | 0.849 |
| RCF-MS [10] | 0.811 | 0.830 | 0.846 |
| RCF-Resnet50 [10] | 0.814 | 0.833 | - |
| CED [11] | 0.816 | 0.833 | **0.889** |
| TB [15] | 0.815 | 0.834 | - |
| DRC | 0.802 | 0.818 | 0.800 |
| DRC-MS-VOC | 0.817 | 0.832 | 0.836 |
| DRC-Resnet50-MS-VOC | 0.819 | 0.835 | 0.850 |
| DRC-Resnet101-MS-VOC | **0.821** | 0.837 | 0.855 |
| DRC-Resnet101-MS-VOC-UCM | 0.819 | **0.843** | 0.505 |

**Network and more training data.** Experiments were conducted with several settings, including the original DRC as a baseline, and training the DRC without weight convolution (W. Conv). Table 3 shows that the proposed weight convolution layer outperforms the original convolution, as the weight convolution can adaptively balance features with different resolutions through weighing coefficient learning. Compared with the single-scale version, the multi-scale DRC (DRC-MS) improved performance. After simply augmenting the standard BSDS500 training dataset with the VOC images as in previous studies [9-11, 15], the proposed method showed better results with ODS-C and ODS-L, and worse results with ODS-T. Therefore, to improve the correctness and localness of the output, additional VOC images are required for training and the contour map should be determined using the multi-scale method. In contrast, to obtain better thickness of the edges without any post-processing, the original DRC should be used. When using Resnet50 and Resnet101 instead of VGG16 as the backbone, the ODS-C and ODS-L performances improved further.

**Efficiency analysis.** Contour detection needs to be computationally efficient to be used as a preprocessing step for high-level applications. Here, the single-scale version achieved a value of 30 FPS (0.041 s) when processing a 481×321 resolution image on a GTX 1080Ti GPU. Because our method is simple, effective, and extremely fast, it is suitable for high-level vision tasks, such as image segmentation. The multi-scale version is slower, with a value of 5.5 FPS (0.182 s) for the same settings.

### 4.4 Comparison with state-of-the-art methods

**BSDS500.** We first compared our method with several non-deep-learning algorithms, such as Canny [19], gPb-UCM [12], and SE [31], and recent deep learning based approaches, such as DeepContour [8], DeepEdge [33], HED [9], CED [11], COB [36], and RCF-MS [10] using boundaries ($F_b$ [23]) evaluation. The left part of Fig. 5 shows the Precision-Recall curves, while Table 4 presents the evaluation results in terms of standard ODS, OIS, and AP. Our DRC achieved better results than the top-performing method in ODS and OIS. This result also surpassed the human benchmark on the BSDS500 dataset with ODS (0.803). RCF-MS [10], CED [11], and TB [15] mixed PASCAL VOC Context into BSDS500, whereas the other methods only used the BSDS training set. The best performance was exhibited by CED [11]. This method achieves better AP performance, as it uses HED pretrained models, apart from ImageNet, on the VGG16 backbone.

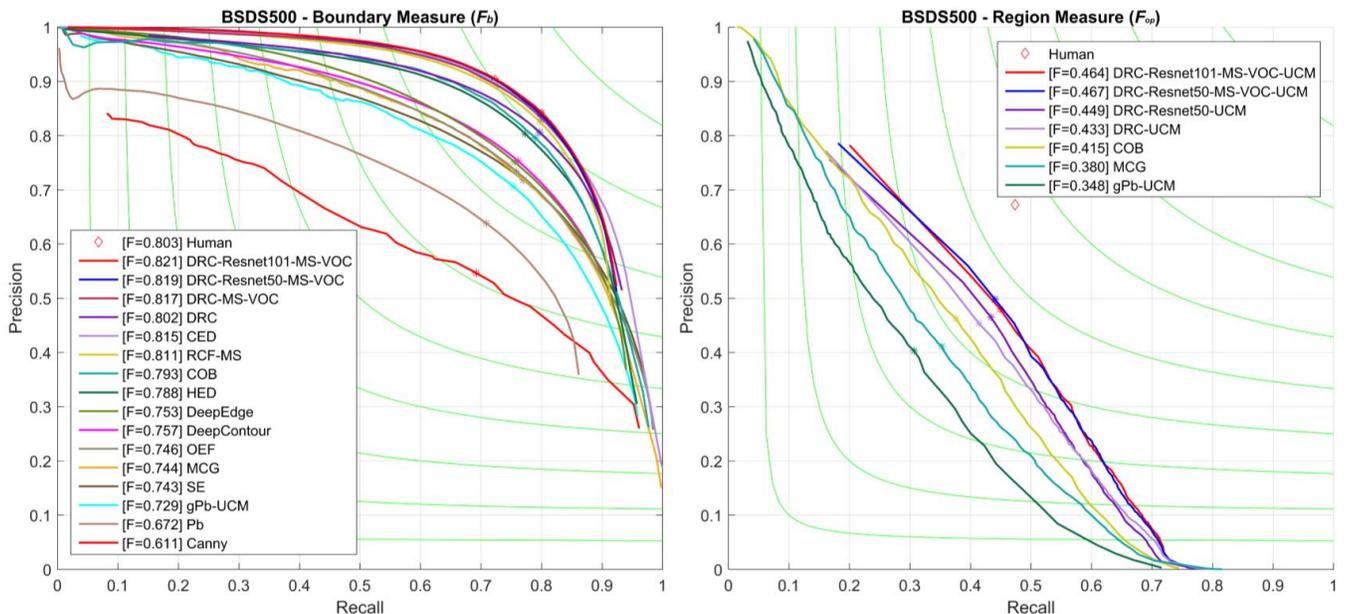

Fig. 5. Precision-recall curves for evaluation of boundaries ($F_b$ [23]), and regions ($F_{op}$ [47]) on BSDS500.

Table 5. Quantitative comparison of boundaries ($F_{op}$ [47]) on BSDS500 dataset.

| Method | ODS | OIS | AP |
|---|---|---|---|
| gPb-UCM [12] | 0.348 | 0.385 | 0.235 |
| MCG [46] | 0.380 | 0.433 | 0.271 |
| COB [36] | 0.415 | 0.466 | **0.333** |
| RCF-Resnet50 [10] | 0.441 | 0.500 | - |
| RCF-Resnet101 [10] | 0.440 | 0.501 | - |
| DRC-UCM | 0.433 | 0.488 | 0.225 |
| DRC-Resnet50-UCM | 0.449 | 0.492 | 0.231 |
| DRC-Resnet50-MS-VOC-UCM | **0.467** | 0.510 | 0.240 |
| DRC-Resnet101-MS-VOC-UCM | 0.464 | **0.512** | 0.225 |

Table 6. Comparison of crispness of edges on BSDS500 dataset.

| Method | ODS-L | OIS-L | ODS-T | OIS-T |
|---|---|---|---|---|
| Human | 0.625 | - | - | - |
| HED [9] | 0.551 | 0.561 | 0.585 | 0.599 |
| CED [11] | 0.609 | 0.619 | 0.655 | 0.662 |
| TB [15] | - | - | 0.693 | 0.700 |
| DRC | 0.601 | 0.610 | **0.697** | **0.705** |
| DRC-MS-VOC | 0.623 | 0.633 | 0.644 | 0.654 |
| DRC-Resnet50-MS-VOC | **0.631** | 0.640 | 0.655 | 0.666 |
| DRC-Resnet101-MS-VOC | **0.631** | 0.641 | 0.657 | 0.667 |

To demonstrate the performances of image segmentation of the proposed method, the edges of DRC were evaluated using an ultrametric contour map (UCM) [12, 46], which can generate different image partitions when thresholding this hierarchical contour map at various contour probability values. Note that MCG [46] needs edge orientations as input. We used an edge map to estimate orientations rather than learning the orientation edge, which may have resulted in a small decrease in performance.

To evaluate the region similarity, the evaluation metric of precision-recall for objects and parts ($F_{op}$) [47] was used. We compared our method with some state-of-the-art methods, such as gPb-UCM [12], MCG [46], COB [36] and RCF [10]. Fig. 5 and Table 5 show that the performance of proposed DRC achieved the new state of the art on BSDS500 region quality. For the region measure, the ODS and OIS F-measure of DRC-Resnet50-MS-VOC-UCM were 6.1% and 2% higher than RCF-Resnet50, respectively. Although the boundary measure reached human performance, the region measure was still far from this standard, indicating that better perception regions should be the main pursuit of classical image segmentation in the future.

We further benchmarked the crispness of edges. We compared our method with the state-of-the-art HED, CED, and the method proposed in [15] (named thick boundary, TB). The results are summarized in Table 6. Our DRC-Resnet101-MS-VOC outperformed the state-of-the-art HED and CED, and achieved ODS-L=**0.631** at $d_0/4$, which surpassed human level performance (ODS-L = 0.625). In addition, the proposed single-scale version of DRC achieved ODS-T=0.697, which slightly exceeds that of the state-of-the-art TB (ODS-T=0.693).

**NYUD.** Fig. 6 shows an overview of the data with the HHA features used in this study, along with the results obtained by DRC. The CNNs were training in three ways: using only RGB data (DRC-RGB); (b) using only HHA data (DRC- HHA); and (c) directly adding RGB and HHA output (DRC-RGB-HHA). The RGB-HHA version obtained significantly better results in standard correctness than the one trained on RGB or HHA data.

We compared our method with several non-deep-learning algorithms, such as gPb-UCM [12], SE [31], gPb+NG [44], SE+NG+ [45] and recent deep learning based approaches, such as HED [9], COB [36], and RCF [10]. The precision-recall curves of boundaries ($F_b$ [23]) are shown in Fig. 7, and the quantitative comparison is presented in Table 7. The best result for the NYUD dataset was achieved by COB [36], where the authors used the more powerful UCM [12] algorithm in post-processing, which has superior for contour performance. We named the version without UCM as COB-original. The proposed DRC was higher than COB-original in standard measurements.

We use the evaluation metric of precision-recall for objects and parts ($F_{op}$) [47]. We compared our method with MCG [46] and COB [36]. As shown in Fig. 7 and Table 8, the performance of the proposed DRC slightly surpass COB in ODS measure of NYUD region quality.

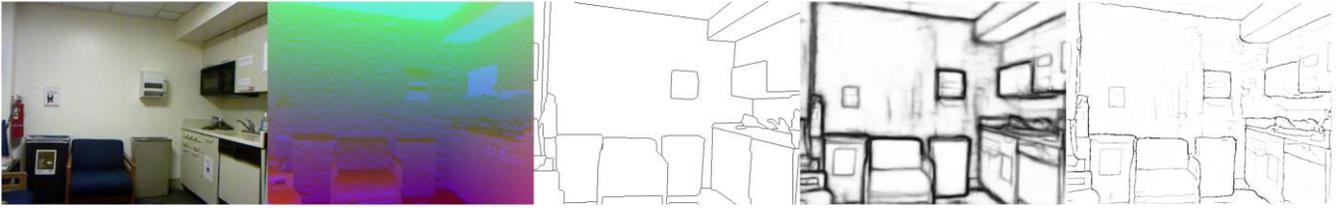

Fig. 6. Data and results on NYUD-v2. From left to right: RGB image, HHA features, ground truth, CNN output, and CNN output with NMS.

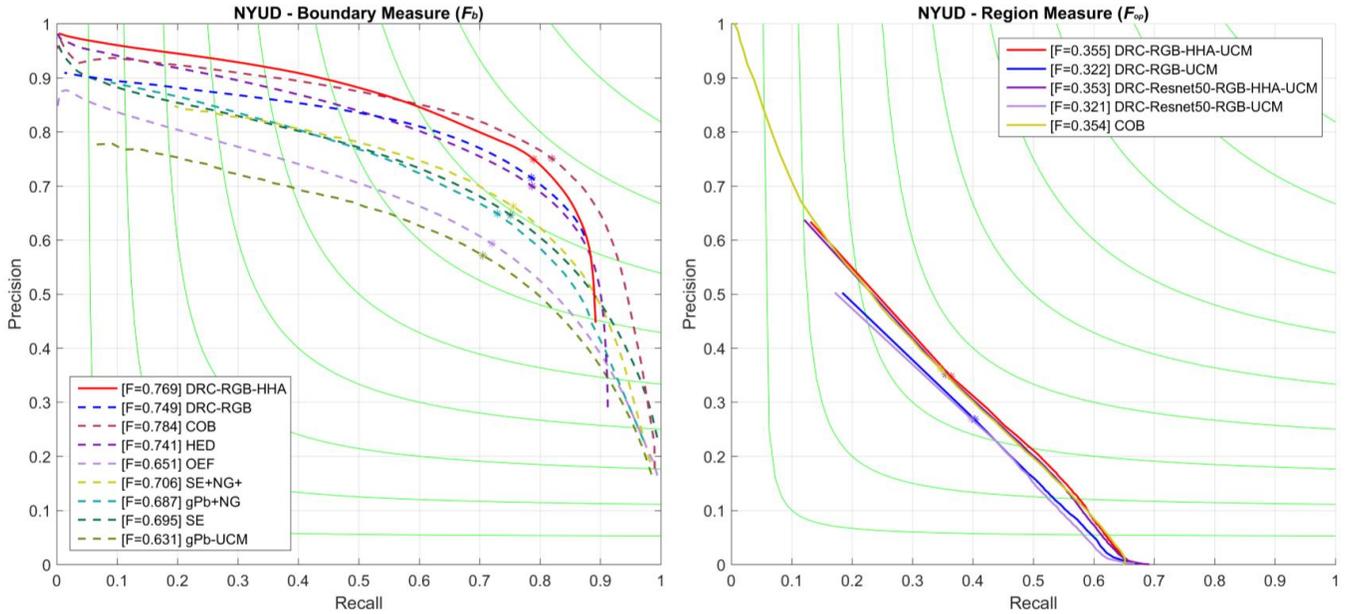

Fig. 7. Precision-recall curves for evaluation of boundaries ($F_b$ [23]), and regions ($F_{op}$ [47]) on NYUD-V2.

Table 7. Quantitative comparison of boundaries ($F_b$ [23]) on NYUD-V2 dataset.

| Method | ODS | OIS | AP |
|---|---|---|---|
| gPb-UCM [12] | 0.631 | 0.661 | 0.562 |
| SE [31] | 0.695 | 0.708 | 0.719 |
| gPb+NG [44] | 0.687 | 0.716 | 0.629 |
| SE+NG+ [45] | 0.706 | 0.734 | 0.549 |
| HED [9] | 0.741 | 0.757 | 0.749 |
| RCF [10] | 0.757 | 0.771 | 0.749 |
| TB [15] | 0.762 | 0.778 | - |
| COB [36] | **0.784** | **0.805** | **0.825** |
| COB-original [36] | 0.745 | 0.762 | 0.792 |
| DRC-RGB | 0.749 | 0.762 | 0.718 |
| DRC-HHA | 0.711 | 0.722 | 0.677 |
| DRC-RGB-HHA | 0.769 | 0.782 | 0.771 |
| DRC-Resnet50-RGB | 0.742 | 0.754 | 0.697 |
| DRC-Resnet50-RGB-HHA | 0.765 | 0.779 | 0.766 |

Table 8. Quantitative comparison of boundaries ($F_{op}$ [47]) on BSDS500 dataset.

| Method | ODS | OIS | AP |
|---|---|---|---|
| MCG [46] | 0.263 | 0.296 | 0.159 |
| COB [36] | 0.354 | **0.397** | **0.267** |
| DRC-RGB-UCM | 0.322 | 0.352 | 0.116 |
| DRC-RGB-HHA-UCM | **0.355** | 0.379 | 0.167 |
| DRC-Resnet50-RGB-UCM | 0.742 | 0.754 | 0.697 |
| DRC-Resnet50-RGB-HHA-UCM | 0.765 | 0.779 | 0.766 |

We further reported the "crispness" of the edges of the proposed DRC. In the NYUD dataset, the default setting of *maxDist*=0.011 was used rather than 0.0075 used for BSDS dataset. Thus, we set *maxDist*=0.011/4=0.00275 for the ODS-L measurement. Table 9 presents the experimental results. The contour accuracy was poor for complex scenes of high resolution. Note that the thickness performance of using Resnet50 had a significant improvement compared with using VGG16. In contrast to the BSDS500 dataset, NYUD-v2 includes high-resolution images, which contain more complex coarse and fine contours. Although DRC obtained a good "crispness" boundary on BSDS500, our results were not quite satisfactory for more complex databases, such as NYUD-v2. Hence, improving "crispness" boundaries should be a focus of future studies.

Table 9. Crispness of edges on NYUD-V2 dataset.

| Method | ODS-L | OIS-L | ODS-T | OIS-T |
|---|---|---|---|---|
| DRC-HHA | 0.288 | 0.293 | 0.370 | 0.382 |
| DRC-RGB | 0.356 | 0.364 | 0.403 | 0.416 |
| DRC-HHA-RGB | **0.359** | **0.368** | 0.403 | 0.436 |
| DRC-Resnet50-RGB | 0.347 | 0.354 | **0.457** | **0.469** |
| DRC-Resnet50-RGB-HHA | 0.354 | 0.362 | 0.423 | 0.434 |

**MultiCue.** Our statistical results indicated single person annotations accounts for 90% of all annotations. Hence, we mapped the SCE weighting $W$ from (0, 1] to (0.5, 1] for best training. We used VGG16 as the backbone. The evaluation results are summarized in Table 10. DRC achieved substantially higher results of than HED and RCF. For boundary tasks, DRC-MS had an F-measure 1.5% or 0.7% higher for ODS or OIS, respectively, than RCF. For edge tasks, DRC-MS had an F-measure 1.1% or 1.0% higher for ODS or OIS, respectively, than RCF. The AP was smaller than HED, probably due to the deep supervision training.

Table 10. Quantitative comparison of performance with the MultiCue dataset.

| Method | ODS | OIS | AP |
|---|---|---|---|
| Human-Boundary [42] | 0.760 (0.017) | - | - |
| Multicue-Boundary [42] | 0.720 (0.014) | - | - |
| HED-Boundary [9] | 0.814 (0.011) | 0.822 (0.008) | **0.869 (0.015)** |
| RCF-Boundary [10] | 0.817 (0.004) | 0.825 (0.005) | - |
| RCF-MS-Boundary [10] | 0.825 (0.008) | 0.836 (0.007) | - |
| DRC-Boundary | 0.820 (0.006) | 0.820 (0.005) | 0.710 (0.006) |
| DRC-MS-Boundary | **0.837 (0.001)** | **0.842 (0.002)** | 0.786 (0.005) |
| Human-Edge [42] | 0.750 (0.024) | - | - |
| Multicue-Edge [42] | 0.830 (0.002) | - | - |
| HED- Edge [9] | 0.851 (0.014) | 0.864 (0.011) | - |
| RCF- Edge [10] | 0.857 (0.004) | 0.862 (0.004) | - |
| RCF-MS-Edge [10] | 0.860 (0.005) | 0.864 (0.004) | - |
| DRC- Edge | 0.859 (**0.002**) | 0.862 (**0.001**) | 0.768 (0.010) |
| DRC-MS- Edge | **0.869 (0.002)** | **0.873** (0.002) | **0.868 (0.002)** |

## 5 DISCUSSION AND CONCLUSION

This study demonstrated that crispness boundary performances can be evaluated by the localness and thickness. To improve these two performances, we proposed a novel method for the challenging tasks. The main novelty of proposed model is summarized as follows. (a) A new CNN architecture (DRNet) was developed. Although the network appears simple, it includes an interesting concept of remixing decoding features. Based on this, we can easily design a more complex network for various tasks. In addition, DRNet involves a useful component, namely the weight convolution layer, which can automatically learn the feature weights in end-to-end learning. The experiments demonstrated that this layer improves performance for the three benchmark evaluations. (b) We developed a novel improved loss function, known as the multi-loss combination framework. This framework allows multiple losses (which often exhibit different distinguishing features) to be easily combined and trained without manually assigning weights. Another contribution is the addition of a multi-label version of the traditional loss, such as cross-entropy and Dice coefficient. These new methods aim to fully use all annotation information, and the experiments verified their advantages.

The quantitative results for analysis of the three benchmark datasets (BSDS, NYUD-V2 and MultiCue) and three measurements (correctness, localness, and thickness) demonstrated that the DRC systematically exceeded the state-of-the-art in most comparisons. In the standard correctness comparison, we achieved a new state-of-the-art performance for the BSDS500 (ODS=0.821) and NYUD-V2 (ODS=0.769) using standard post-processing (NMS and morphological thinning). More importantly, we significantly improved the localness performance, achieving ODS=0.631, which is the first method reported to surpass the human level (ODS=0.625). The thickness performance also achieved a new state-of-the-art. In the NYUD-V2 comparison, the DRC outperformed most of the methods on correctness performances, except COB [36] with the UCM algorithm. The crispness results (see Table 9) showed that for high-resolution image datasets and more rigorous evaluation criteria, the existing contour detection algorithms still have a considerable scope for improvement.

A more challenging task is to determine a method that performs well in both correctness and crispness. Our results demonstrated that the loss function, training methods, components of CNNs, and other factors may influence the final performance. It is clear that the multi-scale and VOC hybrid training method has a significant influence. When we used the multi-scale and mixing VOC for training, the performances of ODS-S and ODS-L significantly increased, and that of ODS-T decreased. Hence, algorithms showing excellent performance for a certain metric should not be selected in all cases. Instead, methods that perform well in both correctness and crispness should be selected as standard components for CNN design.

In this study, we analyzed the evaluation criteria for crisp boundaries and proposed a novel method that focuses on crisp contour detection. The novel CNN architecture of DRNet that makes use of the multi-level refinement modules to build deep networks, providing a new approach that can combine the multiple losses and automatically learn the weights of the different loss functions. Results of ablation studies and comparisons suggest that the high-quality contour detector achieved promising performance for the BSDS500, NYUD-v2 and MultiCue datasets. The proposed methods are expected to inspire subsequent research on contour detection, and further contribute to improving other visual tasks.